\documentclass[letterpaper, 10 pt, conference]{ieeeconf}  

\IEEEoverridecommandlockouts                              

\usepackage{xcolor}
\usepackage{caption}
\usepackage{subcaption}
\usepackage{cite}
\usepackage{booktabs}
\usepackage{multirow}
\usepackage{multicol}
\usepackage{algorithm}
\usepackage{algpseudocode}
\usepackage{url}
\usepackage{hyperref}
\newcommand{\eg}{\emph{e.g.}}
\newcommand{\ie}{\emph{i.e.}}
\newcommand{\etal}{\emph{et~al.}}
\algdef{SE}[SUBALG]{Indent}{EndIndent}{}{\algorithmicend\ }%
\algtext*{Indent}
\algtext*{EndIndent}

\usepackage{graphicx}

\newcommand{\STAB}[1]{\begin{tabular}{@{}c@{}}#1\end{tabular}}

\title{\LARGE \bf
What's in the Black Box? \\The False Negative Mechanisms Inside Object Detectors
}

\author{Dimity Miller$^{1, 2}$, Peyman Moghadam$^{2}$, Mark Cox$^{2}$, Matt Wildie$^{2}$ and Raja Jurdak$^{1}$%
\thanks{$^{1}$Authors are with the Queensland University of Technology (QUT), Brisbane, Australia. $^{2}$Authors are with the Robotics and Autonomous Systems, DATA61, CSIRO, Brisbane, QLD, Australia.}%
\thanks{D.M. and P.M. acknowledge continued support from the CSIRO's Machine Learning and Artificial Intelligence Future Science Platform (MLAI FSP). Contact: {\tt\footnotesize d24.miller@qut.edu.au}. Code is available at https://github.com/csiro-robotics/fn\_mechanisms}}

\begin{document}

\maketitle
\thispagestyle{empty}
\pagestyle{empty}

\begin{abstract}
In object detection, false negatives arise when a detector fails to detect a target object. To understand \emph{why} object detectors produce false negatives, we identify five `false negative mechanisms', where each mechanism describes how a specific component inside the detector architecture failed. Focusing on two-stage and one-stage anchor-box object detector architectures, we introduce a framework for quantifying these false negative mechanisms. Using this framework, we investigate why Faster R-CNN and RetinaNet fail to detect objects in benchmark vision datasets and robotics datasets. We show that a detector's false negative mechanisms differ significantly between computer vision benchmark datasets and robotics deployment scenarios. This has implications for the translation of object detectors developed for benchmark datasets to robotics applications. 
\end{abstract}

\section{INTRODUCTION}
Given an image, object detectors identify target objects in terms of \emph{where} they are and \emph{what} they are. Detectors are useful tools for a range of robotics tasks -- from object-centric mapping~\cite{quadricslam2019}, to search-and-rescue missions~\cite{sandino2021drone}, pest control in protected environments~\cite{liu2021csiro}, and automated harvesting~\cite{sa2016deepfruits}. In each of these applications, the detector is one of the first components in a complex system, and its predictions can dictate the decisions made by follow-on components. As such, the detector has the potential to influence the performance of the entire system.

Within the computer vision community, regular advances have produced object detectors that perform increasingly well on benchmark datasets such as COCO \cite{lin2014microsoft}. While these benchmark datasets are challenging in their own right, they do not always capture the conditions encountered by robotic systems, \eg{} poor lighting, motion blur, unusual viewpoints, and unknown objects. This means that the behaviour of an object detector on a benchmark dataset -- how often it fails and why it fails -- does not always reliably indicate its behaviour when deployed into the real world.

\begin{figure}[t!]
    \centering
    \includegraphics[width=.48\textwidth]{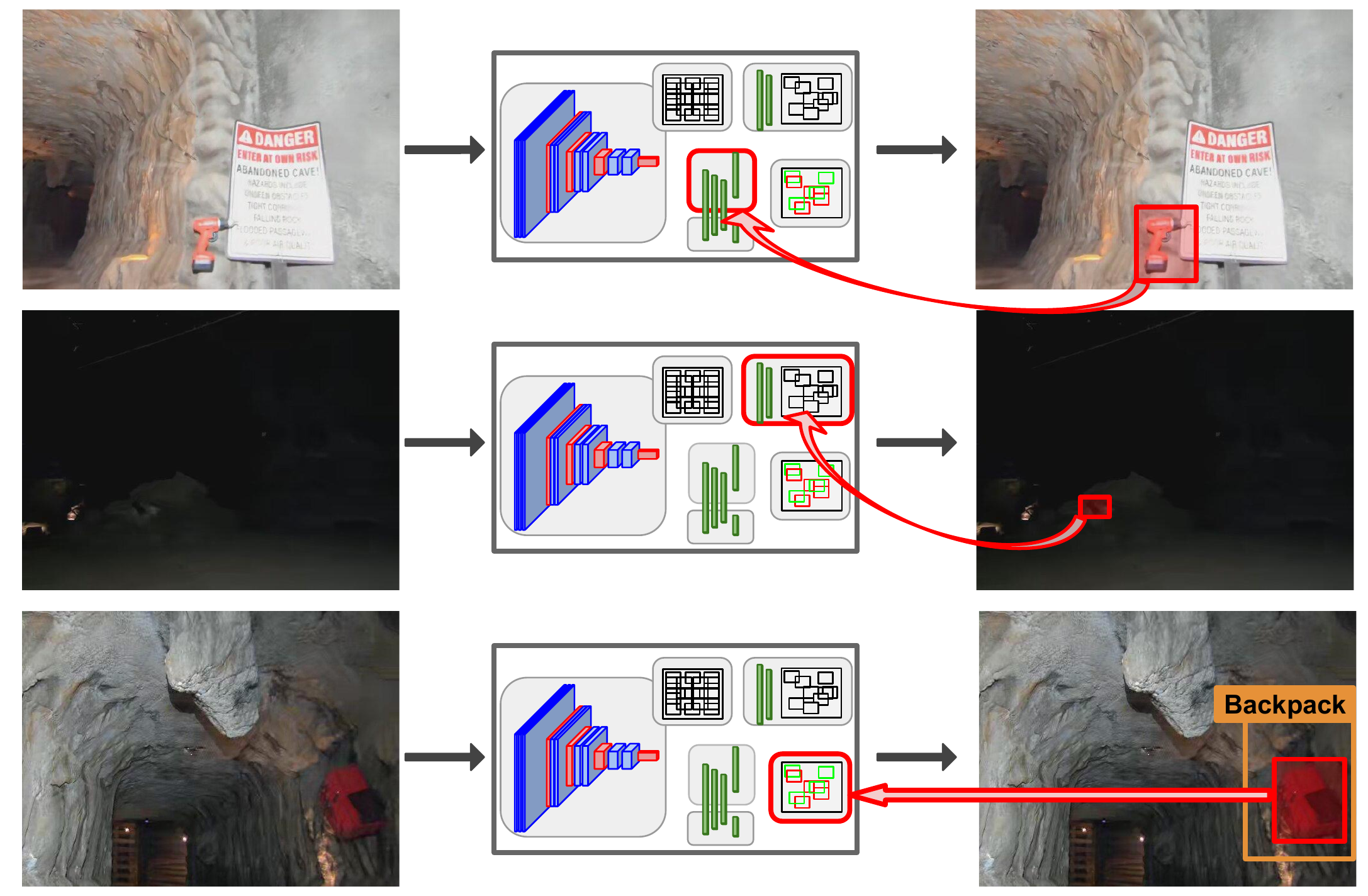}
    \caption{We look inside an object detector to identify which component of the architecture failed to detect an object. In the top image, the classifier confuses the object with the background. In the middle image, the detector's Region Proposal Network doesn't propose a region for the object. In the bottom image, the poorly localised detection suppressed a correctly localised detection. Images are from Team CSIRO Data61 in the DARPA Subterranean Challenge (SubT).  
   }
    \label{fig:front_page}
\end{figure}

In this paper, we focus on understanding why object detectors fail, and how this differs between benchmark datasets and robotics applications. We specifically investigate the phenomenon of false negative failures, where a detector does not detect an object that is present -- either by silently failing to detect the object's presence entirely, or by miscalculating the object's location or classification. We show an example of both in Figure \ref{fig:front_page}. Previous works analysing false negatives describe these errors by examining the output of an object detector, treating the detector itself as a black box \cite{bolya2020tide}. While this provides a descriptive categorisation of detection errors, it does not establish \emph{why} object detectors produce false negatives. To answer this question, we instead examine \emph{inside} the architecture of object detectors, identifying the specific component of the detector architecture that failed.

Focusing on two-stage and one-stage anchor-box object detectors \cite{ren2015faster, lin2017focal, cai2018cascade, redmon2017yolo9000, redmon2018yolov3, liu2016ssd}, we distinguish five false negative mechanisms and introduce a framework for their identification. Using this framework, we investigate different detectors' false negative mechanisms on a combination of benchmark computer vision and robotics datasets that contain challenging conditions. With this approach, we show that the false negative mechanisms of a detector on benchmark datasets do not reliably indicate why a detector will fail in robotics applications with different conditions. In addition, we show that false negative mechanisms from a detector are highly dependent on context-dependent challenging conditions. By taking this alternate approach, we hope to inform future research into how to mitigate false negative errors and improve the reliability of detectors for robotic systems.

\section{BACKGROUND}
\label{sec:bkg}
Object detection is the task of identifying a set of `target' objects in an image. This requires the specification of a set of object categories to be detected, and everything else is considered as part of the `background' of an image and should not be detected. For example, if `person' and `dog' are the only target objects, then trees, fences, frisbees, and everything else, belong to the background category. For a given image, each target object is described by a bounding box that tightly fits the object, and a class label describing the object category. An object detector predicts target objects with a set of detections, where each detection contains a predicted bounding box, a predicted class label, and a score representing confidence in the predicted classification.

On a dataset of images, the primary metric used to summarise object detection performance is mean Average Precision (mAP) \cite{lin2014microsoft, everingham2010pascal}. mAP is the mean of the Average Precision (AP) for each target object class, where AP is the area under a precision-recall curve. The precision-recall curves calculates the number of `true positive' detections, `false positive' detections, and `false negative' objects for each target class. A detection is a true positive if it localises and correctly classifies an object that has not already been detected by a previous detection. Localisation is measured by the Intersection-over-Union (IoU) between the object bounding box and the detection predicted box, and must be greater than a specified threshold -- typically an IoU of at least 0.5 is required \cite{lin2014microsoft, everingham2010pascal}. If the detection does not meet this threshold with any undetected objects of its predicted class, it is a false positive. A false negative describes an object that was not associated with a detection during the evaluation -- \ie{} there were no detections with the correct class label and an IoU greater than 0.5.

\section{PRIOR WORK}
\label{sec:prior}
As established in Section \ref{sec:bkg}, the primary metric assessing object detection performance is mAP. With the mAP metric, failures in object detection can only be coarsely described as false positives or false negatives. To further understand \emph{why} a detection is considered a failure, Hoiem \etal{}~\cite{hoiem2012diagnosing} introduced a categorisation for false positive detections. 

More recently, Bolya \etal{}~\cite{bolya2020tide} built upon this categorisation to include categories describing false negative objects, also improving the generalisability of the categorisation to different datasets. Their work introduces TIDE, which describes six categories for failures in object detectors:

\begin{enumerate}
    \item Classification (Cls) errors: a detection localises an object of the incorrect object class. 
    \item Localisation (Loc) errors: a detection overlaps an object of the correct class, but with poor localisation.
    \item Both Classification and Localisation (Cls+Loc) errors: a detection overlaps an object of the incorrect object class, but with poor localisation.
    \item Duplicate errors: a detection localises an object of the correct object class, but that object has already been associated with a higher scoring detection.
    \item Background errors: a detection does not overlap with an object from any target object class.
    \item Missed GT errors: any undetected ground-truth object that is not due to a Cls, Loc or Cls+Loc error.
\end{enumerate}

These failure categories can represent both false positives and false negatives -- in fact, a false negative failure can be a Cls, Loc, Cls+Loc, or Missed GT error. 

TIDE~\cite{bolya2020tide} is designed specifically as a black box tool for describing object detection errors and analysing their affect on overall detection performance. It categorises false negatives by asking the question -- why didn't the predicted detections capture the false negative object? In contrast, our work asks -- why didn't the detector produce a detection that captured the false negative object? With this distinction, we focus on the detection process occurring \emph{inside} the object detector, and can identify the specific component of the detector that failed and led to the false negative outcome. We refer to these as false negative mechanisms of an object detector.

As we will show in Section \ref{sec:experiments}, despite the utility of TIDE for interpreting and comparing object detection performance, it does not provide a deeper intuitive understanding of the corresponding false negative mechanism. For example, a natural assumption would be that localisation errors are caused by failures in the localisation component of the object detector. This is an \emph{incorrect assumption}. Instead, TIDE localisation errors primarily correspond to failures in the classification component of the detector (\eg{} see Figure \ref{fig:qualimage}). For the purpose of understanding why detectors produce false negatives, our categorisation instead directly highlights the detector component that failed and how.

\begin{figure*}[t!]
    \centering
    \includegraphics[width=.97\textwidth]{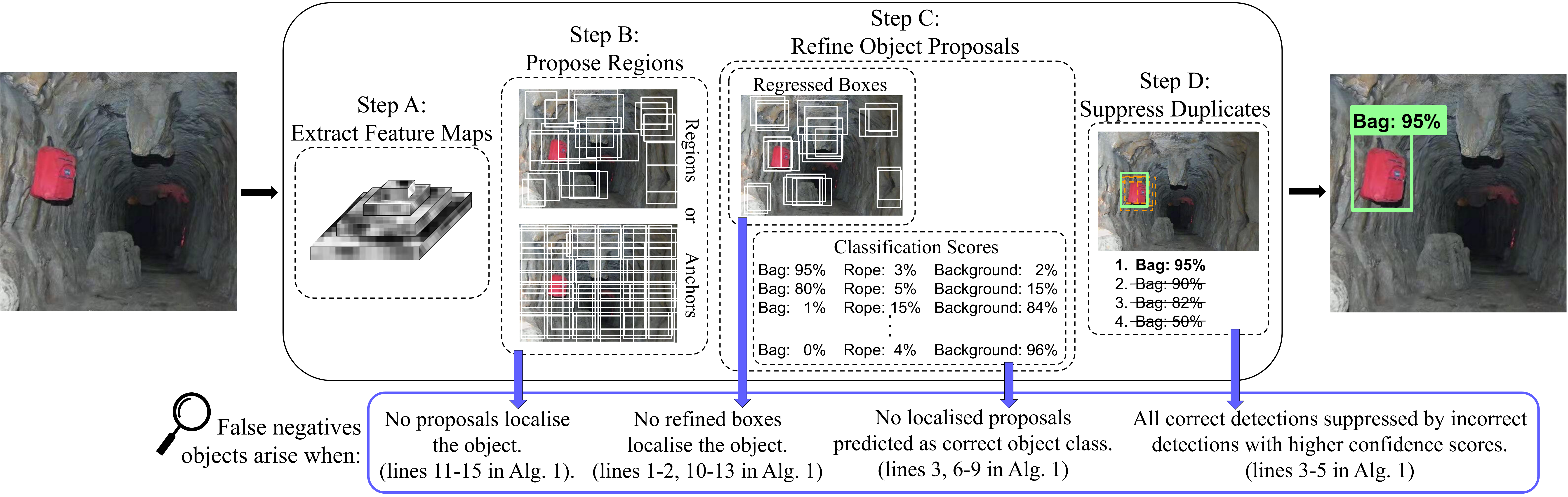}
    \caption{We identify a `detection pipeline' within object detectors that facilitates correct detections of objects. The pipeline contains sequential sections, or `steps', where a failure at any step can prevent an object from being detected and create a false negative -- we refer to these failures in the detection pipeline as `false negative mechanisms'.}
    \label{fig:pipeline}
\end{figure*}

In the last few years, a number of works have emerged introducing techniques that identify when an object detector fails to detect an object~\cite{rahman2019did, ramanagopal2018failing, yang2021introspective}.
These works all highlighted the danger of false negatives in object detectors aboard autonomous systems, arguing potentially catastrophic consequences in applications such as autonomous driving~\cite{rahman2019did, ramanagopal2018failing, yang2021introspective}. Without any existing research informing why object detectors produce false negatives, these works instead focused on exploiting other signals for detecting false negatives -- temporal or stereo camera detection inconsistencies \cite{ramanagopal2018failing}, learning underlying biases in the objects that produce false negatives \cite{yang2021introspective}, and relying on hand-crafted indicators of false negatives in a detector's feature maps \cite{rahman2019did}.

\section{OUR PROPOSED FRAMEWORK}
\label{sec:method}

Our approach for establishing why detectors fail is grounded in understanding how a detector functions to correctly detect an object. In Figure \ref{fig:pipeline}, we visualise a `detection pipeline' used by most two-stage and one-stage anchor-box detector architectures ~\cite{ren2015faster, cai2018cascade, liu2016ssd, redmon2017yolo9000, redmon2018yolov3, lin2017focal}. The pipeline contains four sequential sections, or `steps', where detector components in each step must function correctly for an object to be detected. Below, we describe these steps and then introduce our false negative mechanisms -- ways the detector components can fail and produce a false negative -- and how these can be identified. We formalise this identification of false negative mechanisms in Algorithm \ref{alg:cap}.

\subsection{Extracting Features from an Image}
Before an object can be localised or classified, visual features need to be extracted from image pixels. In object detectors, this is performed by convolutional `backbones', which are typically re-purposed from high-performing image classifiers \cite{lin2017focal, ren2015faster, liu2016ssd, cai2018cascade, redmon2017yolo9000, redmon2018yolov3}. Most modern detectors combine convolutional backbones with a Feature Pyramid Network (FPN) \cite{lin2017focal, cai2018cascade}, which detects objects of various sizes by producing a set of semantically-rich, scaled feature maps \cite{lin2017feature}. 

Feature backbones are ubiquitously initialised with parameters from pre-trained classifiers that perform highly for image recognition \cite{lin2017focal, ren2015faster, liu2016ssd, cai2018cascade, redmon2017yolo9000, redmon2018yolov3}, and are accepted as extracting general image features. While the performance of the backbone may influence down-stream components in the detector architecture, it does not localise or classify objects itself, and thus we do not identify it as a mechanism for false negatives.

\subsection{Proposing Potential Object Regions}
Given the extracted feature maps, the detector identifies image regions that may contain objects. This is handled differently between two-stage and one-stage architectures.

Two-stage architectures rely on fully convolutional networks, known as Region Proposal Networks (RPN) \cite{ren2015faster, cai2018cascade}, to propose a set of image regions that may contain an object. Two-stage detectors initially define a dense grid of `anchors' over the computed feature maps. Each anchor represents a bounding box with a different centre point, scale, and aspect ratio. The RPN is applied to each anchor, predicting `objectness' scores and regressing the anchor boundaries to tightly fit any present object. The anchors with the top-k objectness scores, and their regressed bounding boxes, are then output as the initial set of object proposals.

Similar to two-stage detectors, one-stage anchor-box architectures segment feature maps into a grid of anchor boxes \cite{liu2016ssd, redmon2017yolo9000, redmon2018yolov3, lin2017focal}. However, rather than using an RPN to filter these anchor boxes into region proposals, they instead treat \emph{every} anchor box as a region proposal. This creates a much greater number of object proposals, and instead relies on later steps in the pipeline to filter out the anchors that do not contain an object.

\textbf{Proposal Process False Negative Mechanism:}
We identify that a false negative can be produced if no object proposals localise a target object. We refer to this as a `proposal process' false negative mechanism. Without a proposal for the object, later steps in the detection pipeline will not be input the features necessary to accurately classify or localise the object. We identify this mechanism by computing the IoU between the target object bounding box and all object proposals, where an IoU of at least $\theta_{loc}$ is required for an object to be considered localised (lines 11-15 in Alg. \ref{alg:cap}).  

In a two-stage architecture, the proposal process false negative mechanism may occur for two reasons: (1) none of the RPN regressed anchors sufficiently localised the object, or (2) the anchors that did localise the object were assigned low objectness scores by the RPN.

In a one-stage anchor-box architecture, the proposal process false negative mechanism occurs if no anchor boxes localise a target object. Given the grid-like, dense sampling of anchor boxes, this is far less likely to occur than in a two-stage architecture. For example, given a 640x480 image, a two-stage Faster R-CNN produces 1000 RPN object proposals \cite{ren2015faster}, whereas the one-stage RetinaNet produces 163206 anchor boxes \cite{lin2017focal}.

\subsection{Refining Object Proposals}
The detector then refines the large number of object proposals to determine whether they contain an object, and if so, the classification of the object and its exact location in the image. For both two-stage and one-stage anchor-box detectors, this is typically achieved with two components: (1) a bounding box regressor, and (2) an object classifier. 

\subsubsection{Bounding Box Regressor} To localise the object, regions of interest are extracted from the feature maps for each object proposal bounding box and passed into the bounding box regressor. The bounding box regressor is typically a small Fully Convolutional Network \cite{lin2017focal, ren2015faster, liu2016ssd, redmon2018yolov3, redmon2017yolo9000, cai2018cascade}, and predicts changes, or offsets, to the initially proposed bounding box from the prior step. Often detectors utilise multiple bounding box regressors, with different regressors operating on different feature maps \cite{lin2017focal, liu2016ssd, redmon2017yolo9000, redmon2018yolov3, ren2015faster}, or being used for iterative bounding box refinement \cite{cai2018cascade}.

\textbf{Regressor False Negative Mechanism:} Assuming an object proposal localises an object, the regressor should predict offsets that improve this localisation. However, the regressor may predict poor offsets that de-localise a target object. We refer to this as a `regressor' false negative mechanism. This mechanism is identified when there are no regressed bounding boxes with an IoU of at least $\theta_{loc}$ with the target object, despite the presence of a proposal box that had localised the object (lines 1-2, 10-13 in Alg. \ref{alg:cap}).

\subsubsection{Object Classifier}
Similar to the regressor, the object classifier is typically a small Fully Convolutional Network that intakes the features of each object proposal \cite{lin2017focal, ren2015faster, redmon2017yolo9000, redmon2018yolov3, cai2018cascade, liu2016ssd}. Again, detectors frequently utilise multiple object classifiers to operate on different feature maps \cite{lin2017focal, liu2016ssd, redmon2017yolo9000, redmon2018yolov3, ren2015faster}. In some architectures, the classifier shares layers with the regressor \cite{ren2015faster, redmon2017yolo9000, redmon2018yolov3, liu2016ssd, cai2018cascade}, whereas in others it is an entirely separate network \cite{lin2017focal}. The classifier must distinguish between the different target object classes and the background class. For every object proposal, the classifier outputs a confidence score for each target class and the background class, where high confidence scores indicate the class the proposal belongs to. The YOLO-series detectors \cite{redmon2017yolo9000, redmon2018yolov3} also predict an `objectness' score, which is multiplied with the class confidence scores to create an objectness-aware class confidence score. For all detectors, a minimum confidence score, $\theta_{cls}$, is specified. All proposals with a target class score above this minimum score threshold form a detection, which contains the predicted class label and associated confidence score, and the regressor-refined bounding box. Assuming a proposal has localised an object, the classifier can produce a false negative if it does not assign a classification score above $\theta_{cls}$ to the correct target object class. This can happen for two different reasons, and thus we identify two classification false negative mechanisms at this step in the detection pipeline.

\textbf{Interclass Classification False Negative Mechanism:} In some cases, the classifier can confuse an object for another incorrect target class. We refer to this as an `interclass classification' false negative mechanism. Given the classification scores of boxes that have localised the target object (line 3 in Alg. \ref{alg:cap}), we identify this mechanism when the classifier assigned any incorrect target class a confidence score above $\theta_{cls}$ (lines 6-7 in Alg. \ref{alg:cap}).

\textbf{Background Classification False Negative Mechanism:} The classifier can also fail by misclassifying all proposals of the object as belonging to the background of the image, i.e. the `background' class. We identify this `background classification' mechanism when: (1) a number of boxes localised the object in Step B of the pipeline and (2) none of these localised proposals had any target class confidence scores above $\theta_{cls}$ (lines 3, 6-9 in Alg. \ref{alg:cap}).

\algtext*{EndIf}
\begin{algorithm*}[t!]
\caption{Identifying the Mechanism of a False Negative Object in Object Detection}\label{alg:cap}
\begin{algorithmic}[1]
\small{
\Statex \textbf{Inputs:}
\Indent
\Statex $b, c$ \Comment{the bounding box and class label index of the false negative object}
\Statex $N, N+1$ \Comment{the number of target object classes and the index of the background class}
\Statex $\hat{P} = \{\hat{p_1}, \dots, \hat{p_k}\}$ \Comment{a set of $k$ proposal bounding boxes produced in Step B of the detection pipeline}
\Statex $\hat{B} = \{\hat{b_1}, \dots, \hat{b_k}\}$ \Comment{a set of $k$ bounding boxes refined by the regressor in Step C of the detection pipeline}
\Statex $\hat{\textbf{S}} = \{\hat{S_1}, \dots, \hat{S_k}\}$, \Comment{a set of classification score lists from Step C in the detection pipeline}
\Indent 
\Statex where $\hat{S_i} = [s_{i,1}, \dots, s_{i,N}, s_{i,N+1}]$ \Comment{a list of confidence scores for each target class and the background class}
\EndIndent
\Statex $\theta_{loc}, \theta_{cls}$ \Comment{IoU threshold for object localisation and score threshold for object classification}
\EndIndent
\Statex \textbf{Compute:}
\Indent
\State $IoU_{\hat{B}} = IoU(b, \hat{B})$ \Comment{Calculate IoU between all regressor-refined boxes and the false negative object box}
\If{any $IoU_{\hat{B}} \geq \theta_{loc}$} 
\Comment{Check if the object was correctly localised by a regressed box}
    \State $S_{localised} = \mathbf{S}[IoU_{\hat{B}} \geq \theta_{loc}]$ \Comment{The set of classification score lists for the correctly localised boxes}
    \If{any $S_{localised}[c] \geq \theta_{cls}$} \Comment{Check if any localised boxes had also correctly classified the target object}
    \State \textbf{return} ``Classifier Calibration" \Comment{A correct detection was suppressed in Step D of the pipeline}
    \ElsIf{any $S_{localised}[1:N] \geq \theta_{cls}$} \Comment{Check if any localised boxes were misclassified as another target class}
    \State \textbf{return} ``Interclass Classification"   \Comment{The object was misclassified for another target class in pipeline Step C}
    \Else{} \Comment{All localised boxes were misclassified as the background class}
    \State \textbf{return} ``Background Classification" \Comment{The object was misclassified as the background in pipeline Step C}
    \EndIf
\Else{} \Comment{The object was not localised by any regressed boxes}
\State $IoU_{\hat{P}} = IoU(b, \hat{P})$  \Comment{Calculate IoU between all object proposal boxes and the false negative object box}
\If{any $IoU_{\hat{P}} \geq \theta_{loc}$} \Comment{Check if the object was correctly localised by any object proposal boxes}
    \State\textbf{return} ``Regressor'' \Comment{Any localised object proposals were incorrectly regressed in Step C of the pipeline}
\Else{} \Comment{No object proposals localised the object}
    \State\textbf{return} ``Proposal Process'' \Comment{No proposals localised the object in Step B of the pipeline}
\EndIf
\EndIf
\EndIndent}
\end{algorithmic}
\end{algorithm*}

\subsection{Suppressing Duplicate Detections}
Given the large number of object proposals from Step B, there are often multiple overlapping detections that jointly detect a single object with the same predicted class label. Non-Maximum Suppression (NMS) is an algorithm commonly used by object detectors to suppress duplicate detections of a single object \cite{redmon2017yolo9000, redmon2018yolov3, liu2016ssd, cai2018cascade, ren2015faster, lin2017focal}. For each target class, it ranks all predicted detections from the highest class confidence score to the lowest. Moving from top to bottom, any detection that highly overlaps with a more confident detection is suppressed. Typically, an overlap threshold of IoU greater than 0.5 is used. Any detections remaining are finally output as object predictions from the object detector.

\textbf{Classifier Calibration False Negative Mechanism:} NMS relies on the best localised bounding boxes having the highest classification confidence scores. If this assumption is not satisfied, a false negative can be introduced when a correctly localised detection is suppressed by a detection that has not localised the object, \ie{} IoU less than $\theta_{loc}$. We refer to this as a `classifier calibration' false negative mechanism, as it is ultimately the fault of the calibration of the classifier's confidence scores -- detections that better localise an object should be coupled with higher confidence scores from the classifier. Prior to NMS, we can identify this false negative mechanism when there is a regressed bounding box that localises the target object with an IoU greater than $\theta_{loc}$ (correctly localised) (lines 1-2 in Alg. \ref{alg:cap}), that also has a confidence score for the correct target class above $\theta_{cls}$ (correctly classified) (lines 3-5 in Alg. \ref{alg:cap}), but it was not produced as an output detection. 

\section{EXPERIMENTAL SETUP}
\label{sec:expdesign}
In this section, we detail the object detectors, datasets, and implementation details used to investigate our framework for identifying false negative mechanisms.

\subsection{Object Detectors}
\label{sec:network}
We evaluate with two object detectors: Faster R-CNN \cite{ren2015faster}, and RetinaNet \cite{lin2017focal}. Faster R-CNN is the classic two-stage object detector, and despite being developed nearly 7 years ago, it maintains competitive performance on the COCO benchmark challenge when combined with state-of-the-art feature backbones \cite{wu2019detectron2}. RetinaNet is a popular one-stage anchor-box object detector, and the first of its kind to obtain competitive performance with two-stage detectors \cite{lin2017focal}.

\subsection{Datasets}
\label{sec:dataset}
We evaluate across a combination of benchmark computer vision and robot-collected datasets:

The \textbf{COCO dataset}~\cite{lin2014microsoft} is the predominant benchmark dataset for evaluating object detectors. We evaluate with the `val2017' split of the dataset, which contains 4952 images featuring objects from 80 different target classes.
\\
The \textbf{Exclusively Dark (ExDark) dataset}~\cite{loh2019getting} is a public dataset containing 7363 images from low-light indoor and outdoor environments. Target object classes include bicycle, boat, bottle, bus, car, cat, chair, cup, dog, motorbike, person, and table.  While challenging illumination is relevant to robotics applications, ExDark was collated with images sampled from internet websites rather than robot-collected.
\\
The \textbf{iCubWorld Transformation (iCubWorldT) dataset} \cite{pasquale2019we} is a public dataset collected by an iCub humanoid robot. Images were collected with a human randomly rotating an object before the robot. As a result, the dataset features objects from a variety of challenging viewpoints. We test on 2400 images that contained target object classes overlapping with COCO, specifically images of books, bottles, and cups.
\\
The \textbf{CSIRO Data61 Subterranean Robot dataset} consists of 5384 images collected by Team CSIRO Data61 \cite{hudson2021heterogeneous} during the final event of the DARPA Subterranean (SubT) Challenge \cite{subtchallenge}. The challenge required a robot fleet to explore an unknown subterranean environment under the supervision of a single human operator, and detect and localise a set of target objects -- namely survivor, backpack, fire extinguisher, helmet, rope, cell phone, colored cube, drill, and air-vent. The test images in this dataset are from the RGB cameras aboard Team CSIRO Data61's robot fleet, and were collated by extracting frames with tracked artefacts from the team's 3D tracking and artefact visibility system. For each extracted frame, we hand-labelled annotations for any present objects.

\begin{table}[t]
    \centering
    \caption{False negative rates on each dataset.}
    \label{tab:FN_Nums}
    \begin{tabular}{@{}lrrr@{}}
         \toprule
          & \textbf{\# Objects} & \multicolumn{2}{c}{\textbf{False Negative Rate (\#)}}\\ 
          & & \textbf{Faster R-CNN} & \textbf{RetinaNet} \\
         \midrule
         COCO \cite{lin2014microsoft} & 36335 & 28.8\% (10464) & 32.7\% (11869)\\
         iCubWorld \cite{pasquale2019we}& 2400 & 37.5\% (901) & 48.5\% (1165)\\
         ExDark \cite{loh2019getting} & 23710 & 28.4\% (6737) & 30.3\% (7181) \\
         CSIRO Data61 SubT & 5428 & 17.2\% (936) & 16.5\% (895) \\
         \bottomrule
    \end{tabular}
\end{table}

\subsection{Implementation}
For both detectors, we use the publicly available {\fontfamily{qcr}\selectfont
detectron2} implementation \cite{wu2019detectron2} and a ResNet50 \cite{he2016deep} with Feature Pyramid Network \cite{lin2017feature} backbone. For experiments on COCO, ExDark and iCubWorldT, the detectors are trained with the `train' subset of the COCO dataset \cite{lin2014microsoft}. For experiments on the CSIRO Data61 SubT dataset, we train both detectors with the training dataset used by Team CSIRO Data61, which consists of approximately 28,000 images collected in-house. When implementing the algorithm for identifying false negative mechanisms, we use the standard localisation threshold $\theta_{loc}$ of $0.5$~\cite{lin2014microsoft, everingham2010pascal}, and a classification score threshold $\theta_{cls}$ of $0.3$.

\section{RESULTS}
\label{sec:experiments}

\subsection{False Negative Mechanisms versus Error Types}
\begin{figure*}[t!]
    \centering
    \includegraphics[width=.84\textwidth]{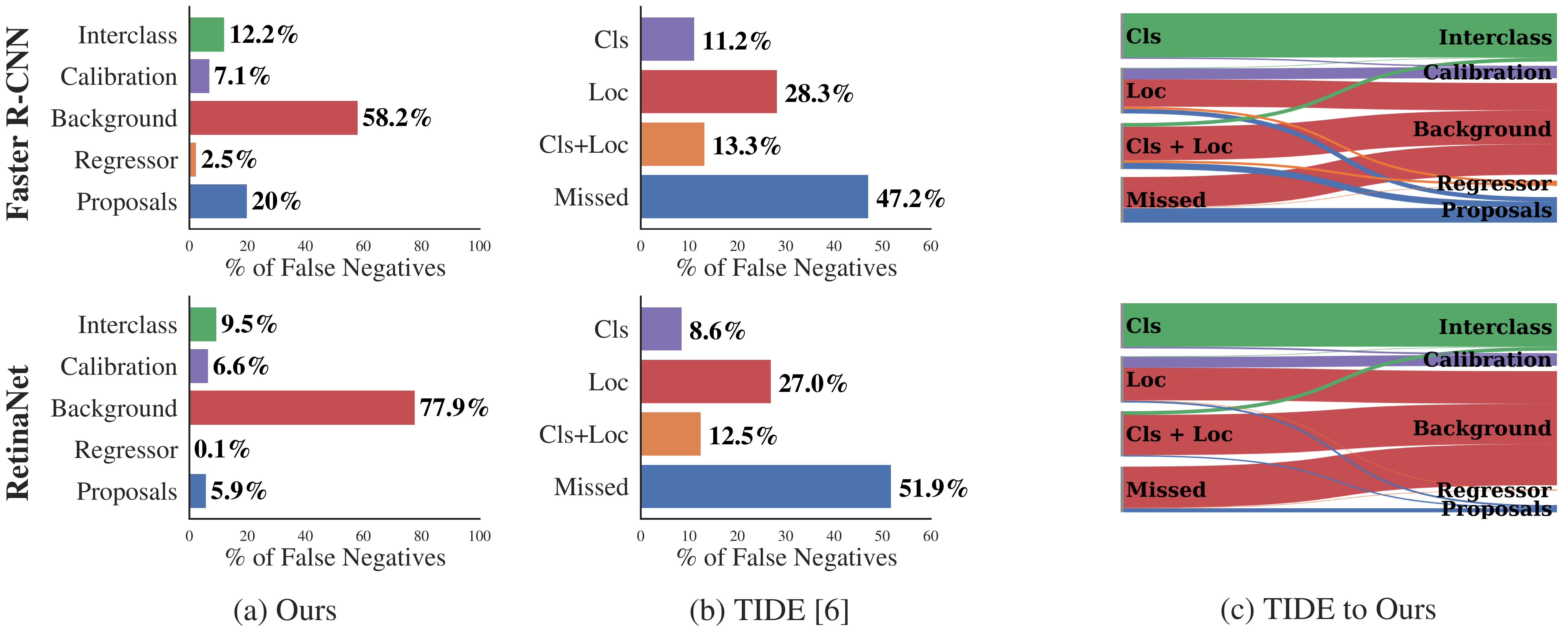}
    \caption{We compare (a) our identified false negative mechanisms to (b) the TIDE \cite{bolya2020tide} error types when detectors are tested on COCO val2017. In (c), a Sankey plot visualises the mapping from TIDE error types (left) to our identified false negative mechanisms (right), where the width of the connecting lines is proportionate to the relative composition of each error type.}
    \label{fig:cocoresults}
\end{figure*}

\begin{figure*}[t!]
    \centering
    \includegraphics[width=.8\textwidth]{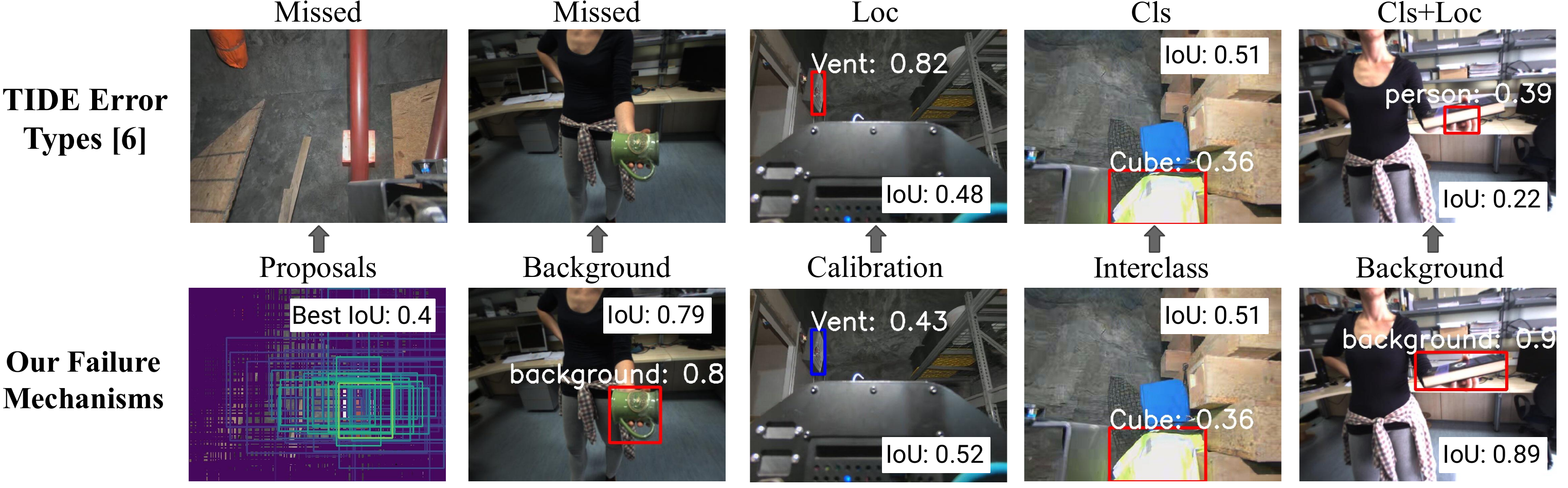}
    \caption{For a selection of Faster R-CNN false negatives, we compare the TIDE error type \cite{bolya2020tide} with our identified failure mechanism. The IoU with the target object is indicated where relevant.}
    \label{fig:qualimage}
\end{figure*}

\begin{figure*}[t!]
    \centering
    \includegraphics[width=.84\textwidth]{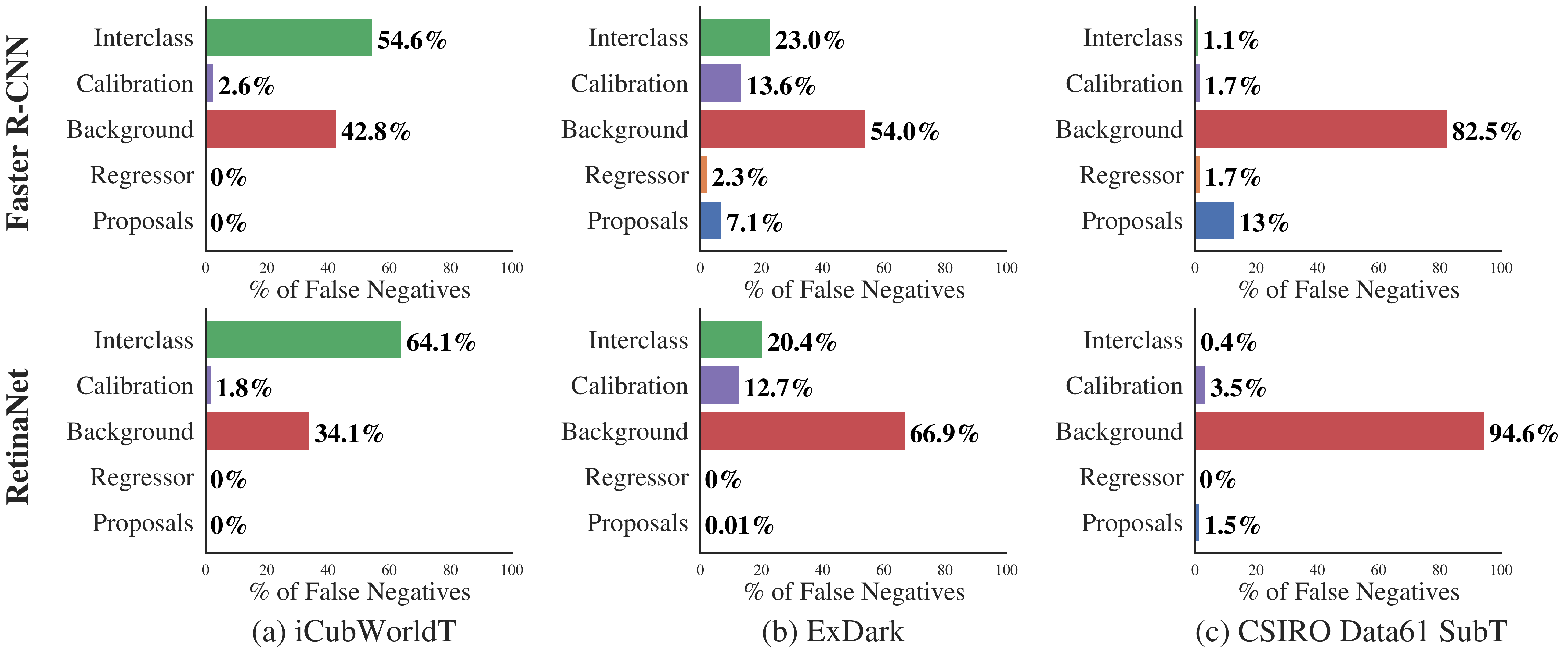}
    \caption{The distribution of our identified false negative mechanisms on datasets containing challenging perception conditions.}
    \label{fig:challengingFN}
\end{figure*}

As shown in Table \ref{tab:FN_Nums}, Faster R-CNN and RetinaNet fail to detect 28.8\% and 32.7\% of all target objects in COCO \cite{lin2014microsoft}. In Figure \ref{fig:cocoresults}, from left to right, we compare the distribution of our identified false negative mechanisms, the distribution of TIDE's false negative error types \cite{bolya2020tide}, and the relationship between our false negative mechanisms and the TIDE error types. 
Examining our identified false negative mechanisms, two key insights are clear across both detector architectures: 
\begin{enumerate}
    \item The classifier causes most false negatives -- between 77.5\% (Faster R-CNN) to 94\% (RetinaNet).
    \item The regressor rarely fails to localise an object -- between 0.1\% (RetinaNet) to 2.5\% (Faster R-CNN) of all false negatives.
\end{enumerate}

The Background Classification false negative mechanism is the primary failure mechanism, where a bounding box that localised the object existed, but was misclassified as belonging to the background class. Given the well-established challenge of foreground-background class imbalance in object detection \cite{oksuz2020imbalance}, this result is consistent with existing literature. 

RetinaNet particularly struggles with this foreground-background imbalance, with 19.7\% more Background Classification mechanisms than Faster R-CNN. Instead, Faster R-CNN exhibits an 14.1\% increase in Proposal Process false negatives compared to RetinaNet. These differences are again consistent with the expected challenges of each architecture -- Faster R-CNN first filters out anchor boxes of the image background with its Region Proposal Network, and then again later with the classifier~\cite{ren2015faster}. In contrast, RetinaNet does not filter anchor boxes during the proposal process, relying solely on the classification component of the architecture to filter out the large number of background proposals~\cite{lin2017focal}.  

Examining the false negative error types introduced by the TIDE categorisation \cite{bolya2020tide}, the majority of false negatives belong to the `Missed' category -- between 47.2\% (Faster R-CNN) and 51.9\% (RetinaNet). The remaining false negatives mostly belong to a `Localisation' error category (Localisation alone, or Localisation and Classification) -- 39.5\% (RetinaNet) to 41.6\% (Faster R-CNN). 

The right side of Figure \ref{fig:cocoresults} shows the composition of our identified false negative mechanisms for each of the TIDE error types. We observe that most TIDE error types can be produced by multiple different false negative mechanisms. For example, in Figure \ref{fig:qualimage}, while the `Missed' error types appear identical from the output of the detector, looking inside the detector can reveal either a Proposal Process or Background Classification false negative mechanism. Additionally, TIDE error types do not always intuitively signal the underlying the false negative mechanism. For example, while the `Localisation' error type suggests a failure in the localisation component of an object detector, the majority of these false negatives are caused by classifier mechanisms. An example is shown in Figure~\ref{fig:qualimage}, where the `Localisation' error was caused by a `Classifier Calibration' mechanism -- a correctly localised detection existed, but was suppressed due to a low confidence score. Notably, TIDE \cite{bolya2020tide} remains a useful tool for comparing the performance of different object detectors. However, our framework and proposed false negative mechanisms are necessary to perform a deeper analysis and clearly isolate the specific detector component that produced a false negative.

\subsection{False Negative Mechanisms in Challenging Conditions}
Table \ref{tab:FN_Nums} and Figure \ref{fig:challengingFN} show respectively the false negative rate and distribution of our identified false negative mechanisms on the datasets with challenging perception conditions. We observe three key insights from these results:

\begin{enumerate}
    \item The distribution of false negative mechanisms is fairly consistent amongst the internet-collected, benchmark datasets COCO and ExDark -- yet this distribution does not reliably indicate the false negative mechanisms present on the robot-collected datasets iCubWorldT and CSIRO Data61 SubT.
    \item Among the robot-collected datasets, discrepant false negative mechanisms are observed. For iCubWorldT, the predominant mechanism is Interclass Classification, whereas for CSIRO Data61 SubT, the predominant mechanism is Background Classification.
    \item Despite the differences across the datasets, the distribution of false negative mechanisms is consistent across the two different detector architectures.
\end{enumerate}

The iCubWorldT dataset, featuring challenging viewpoints of objects in indoor environments, primarily produces false negatives through the Interclass Classification mechanism -- an increase of 44.2\% (Faster R-CNN) and 54.6\% (RetinaNet) compared to COCO. This indicates that the primary challenge for objects in unusual viewpoints is discriminating between target object classes. We infer this is due to a lack of training images with those viewpoints, thus the detector cannot learn to discriminate at those viewpoints.

For the CSIRO Data61 SubT dataset, the Background Classification mechanism is the dominant source of false negatives, responsible for 82.8\% (Faster R-CNN) to 94.6\% (RetinaNet) of all false negatives. All other false negative mechanisms are significantly reduced or negligible. The difference in false negative mechanisms across the standard benchmark datasets and the robot-collected datasets has interesting implications about the transfer of research from the computer vision community to the robotics community -- techniques developed to improve detectors for benchmark datasets may not offer the same performance improvement in a robotics application. For example, while ``solving'' all non-Background mechanisms would yield a 22.1\% (RetinaNet) to 41.2\% (Faster R-CNN) reduction in false negatives on COCO, it would only remove 5.4\% (RetinaNet) to 17.5\% (Faster R-CNN) of false negatives for the subterranean application represented by CSIRO Data61 SubT. 

ExDark follows the false negative mechanism trends of the COCO dataset. While there are slight differences -- less Background Classification and Proposal Process mechanisms and more Interclass Classification and Classifier Calibration mechanisms -- the distributions are mostly consistent. We infer this similarity is because ExDark is composed of internet-collated images, similar to COCO, and thus is likely to contain similar biases in image structure and content. 

\subsection{Investigating the Influence of the Detector Backbone}
Table \ref{tab:backbone} shows the effect of different backbone networks -- ResNet50, ResNet101~\cite{he2016deep}, and ResNeXt101~\cite{Xie_2017_CVPR} -- on the detector false negative mechanisms and false negative rate. For both detectors, across COCO and ExDark, the backbone has limited effect on the distribution of false negative mechanisms produced. This suggests that the false negative mechanisms of a detector are mostly independent of the backbone, and instead are a function of the detector architecture or training paradigm.

\begin{table}[t]
    \centering
    \caption{The influence of the backbone (BB) on detector false negative rate (FNR) and false negative mechanisms.}
    \label{tab:backbone}
    \begin{tabular}{@{}cclcccccc@{}}
         \toprule
         && \textbf{BB} & \textbf{FNR} & \multicolumn{5}{c}{\textbf{FN Mechanism Distribution}} \\
         &&& & Prop & Reg & Bkg & Cal & Inter \\
         \midrule
         \multirow{6}{*}{\STAB{\rotatebox[origin=c]{90}{Faster R-CNN}}}
         & \multirow{3}{*}{\STAB{\rotatebox[origin=c]{90}{COCO}}}
         & R50 & 30.0\% & 20.0\% & 2.5\% & 58.2\% & 7.1\% & 12.2\% \\
         & & R101 & 28.2\% & 19.3\% & 2.4\% & 59.0\% & 7.4\% & 11.9\% \\
         & & X101 & 27.6\% & 20.1\% & 2.8\% & 58.9\% & 7.7\% & 10.5\% \\
        \cmidrule[0.5pt](ll){2-9}
        & \multirow{3}{*}{\STAB{\rotatebox[origin=c]{90}{ExDark}}}
         & R50 & 28.4\% & 7.1\% & 2.3\% & 54.0\% & 13.6\% & 23.0\% \\
         & & R101 & 28.3\% & 6.3\% & 2.4\% & 55.5\% & 13.8\% & 22.0\% \\
         & & X101 & 29.1\% & 5.9\% & 2.4\% & 57.8\% & 13.5\% & 20.4\% \\
         \midrule
         \multirow{6}{*}{\STAB{\rotatebox[origin=c]{90}{RetinaNet}}}
         & \multirow{3}{*}{\STAB{\rotatebox[origin=c]{90}{COCO}}}
         & R50 & 32.7\% & 5.9\% & 0.1\% & 77.9\% & 6.6\% & 9.5\% \\
         & & R101 & 32.2\% & 5.8\% & 0.1\% & 78.5\% & 6.9\% & 8.7\% \\
         & & X101 & 30.8\% & 5.8\% & 0.1\% & 78.0\% & 7.4\% & 8.7\% \\
         \cmidrule[0.5pt](ll){2-9}
         & \multirow{3}{*}{\STAB{\rotatebox[origin=c]{90}{ExDark}}}
         & R50 & 30.3\% & 0.0\% & 0.0\% & 66.9\% & 12.7\% & 20.4\% \\
         & & R101 & 30.8\% & 0.0\% & 0.0\% & 67.6\% & 12.8\% & 19.6\% \\
         & & X101 & 29.8\% & 0.0\% & 0.0\% & 68.4\% & 12.7\% & 18.9\% \\
         \bottomrule
    \end{tabular}
\end{table}

\subsection{Discussion on Mitigating False Negatives}

As investigated in ~\cite{rahman2019did, ramanagopal2018failing, yang2021introspective}, false negatives can be mitigated with additional systems that `warn’ when a detector potentially fails. Our analysis showed that most false negatives arise from Background Classification mechanisms, where the classifier fails to distinguish target objects from the image background. In two-stage architectures such as Faster R-CNN, the Region Proposal Network (RPN) performs a similar task by identifying potential objects with high `objectness’ scores; interestingly, for all Background Classification failures, the RPN performed this task reliably. This suggests that information from a RPN may be a meaningful input into `warning' systems, \textit{i.e.} image areas assigned high `objectness’ by a RPN without any output detection may indicate a potential false negative failure.

False negatives can also be reduced by adapting the architecture or training procedure of object detectors. Our results showed that unusual viewpoints of objects (as in iCubWorldT) elicit high levels of Interclass Classification false negatives. These unusual viewpoints may be underrepresented in the training data, and therefore the detector’s classifier does not learn to distinguish certain object viewpoints. Our proposed framework could be used to inform the collection of targeted data, such as viewpoints that elicit Interclass Classification false negatives, which can be used to retrain a more robust detector. In addition to this, the prevalence of Background Classification false negative mechanisms indicates that `background imbalance’ remains an ongoing challenge for object detectors \cite{oksuz2020imbalance}. Further research addressing this imbalanced learning may contribute to reducing the overall quantity of Background Classification false negatives produced by a detector.

\section{CONCLUSIONS}
\label{sec:conclusions}

In this paper, we identified five mechanisms within one-stage and two-stage anchor-box object detectors that can cause false negative failures. The overarching goal of this work is to enable future research into object detection architectures that are robust to false negatives. Yet quantifying the false negative mechanisms of an object detector also has broader implications. With the ability of robots to localise in their environments, our framework could also be adapted to an online learning scenario -- after detecting an object that was undetected in a previous frame, the system could analyse which component of the detector failed, and perform targeted online learning with any collected data. Future work is also necessary to extend the concept of false negative mechanisms to other architectures, including anchor-free or transformer-based detectors.




\bibliographystyle{IEEEtran}
\bibliography{refs}

\end{document}